\documentclass[10pt,twocolumn,letterpaper]{article}

\usepackage{wacv}              

\usepackage{graphicx}
\usepackage{amsmath}
\usepackage{amssymb}
\usepackage{booktabs}

\usepackage[pagebackref,breaklinks,colorlinks]{hyperref}

\usepackage[capitalize]{cleveref}
\crefname{section}{Sec.}{Secs.}
\Crefname{section}{Section}{Sections}
\Crefname{table}{Table}{Tables}
\crefname{table}{Tab.}{Tabs.}

\def\wacvPaperID{*****} 
\def\confName{WACV}
\def\confYear{2025}

\usepackage{multirow}
\usepackage[flushleft]{threeparttable}
\usepackage{lipsum}

\begin{document}

\title{HSViT: Horizontally Scalable Vision Transformer}

\author{Chenhao Xu\\
School of IT, Deakin University\\
Geelong, VIC, Australia\\
{\tt\small chenhao.xu@deakin.edu.au}
\and
Chang-Tsun Li\\
School of IT, Deakin University\\
Geelong, VIC, Australia\\
{\tt\small changtsun.li@deakin.edu.au}
\and
Chee Peng Lim\\
IISRI, Deakin University\\
Geelong, VIC, Australia\\
{\tt\small chee.lim@deakin.edu.au}
\and
Douglas Creighton\\
IISRI, Deakin University\\
Geelong, VIC, Australia\\
{\tt\small douglas.creighton@deakin.edu.au}
}

\maketitle

\begin{abstract}
Due to its deficiency in prior knowledge (inductive bias), Vision Transformer (ViT) requires pre-training on large-scale datasets to perform well. Moreover, the growing layers and parameters in ViT models impede their applicability to devices with limited computing resources. To mitigate the aforementioned challenges, this paper introduces a novel horizontally scalable vision transformer (HSViT) scheme. Specifically, a novel image-level feature embedding is introduced to ViT, where the preserved inductive bias allows the model to eliminate the need for pre-training while outperforming on small datasets. Besides, a novel horizontally scalable architecture is designed, facilitating collaborative model training and inference across multiple computing devices. 
The experimental results depict that, without pre-training, HSViT achieves up to $10\%$ higher top-1 accuracy than state-of-the-art schemes on small datasets, while providing existing CNN backbones up to $3.1\%$ improvement in top-1 accuracy on ImageNet.
The code is available at \url{https://github.com/xuchenhao001/HSViT}.
\end{abstract}

\section{Introduction}
\label{sec:introduction}

\begin{figure}
    \centering
    \includegraphics[width=\linewidth]{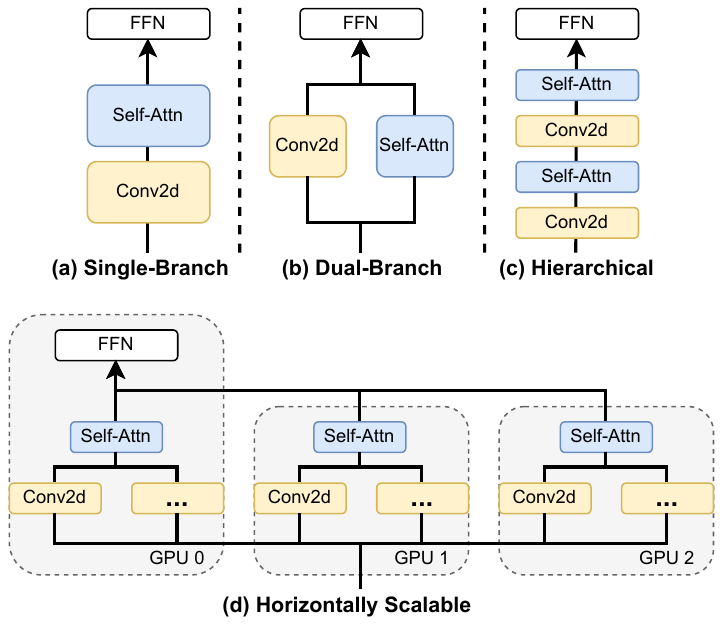}
    \caption{Comparison of various hybrid ViT architectures. HSViT can be deployed across multiple computing devices to utilize resources better.}
    \label{fig:arch_comparison}
\end{figure}

Vision Transformers (ViTs) typically necessitate extensive pre-training on large-scale datasets to achieve their superior performance~\cite{dosovitskiy2021image}. This requirement arises from their lack of inductive biases, which are inherently present in convolutional neural networks (CNNs)~\cite{raghu2021vision, wu2021cvt}. By contrast, ViTs divide images into a sequence of fixed-sized, non-overlapping patches and apply positional encodings to these patches~\cite{dosovitskiy2021image}. This patch-based approach hinders the self-attention layers to capture spatial invariances~\cite{raghu2021vision}. Consequently, hybrid ViTs, which combine self-attention mechanisms with convolutional layers, offer a promising solution for capturing long-range dependencies in image features while preserving inductive biases.

As shown in Fig.~\ref{fig:arch_comparison}, there are primarily three designs of hybrid ViT: (a) single-branch~\cite{carion2020end, chen2020dynamic, dosovitskiy2021image, li2022efficientformer}, (b) dual-branch~\cite{pan2022integration, zhang2022bootstrapping}, and (c) hierarchical~\cite{wu2021cvt, mehta2022mobilevit, guo2022cmt, shaker2023swiftformer}. The single-branch and dual-branch architectures typically integrate full-size CNN and ViT models, while the hierarchical design incorporates multiple self-attention layers into classic CNNs to build a hierarchical feature extraction structure. However, these methods often overlook horizontal scalability, particularly in the context of utilizing multiple computing devices such as GPUs.

\begin{figure}
    \centering
    \includegraphics[width=\linewidth]{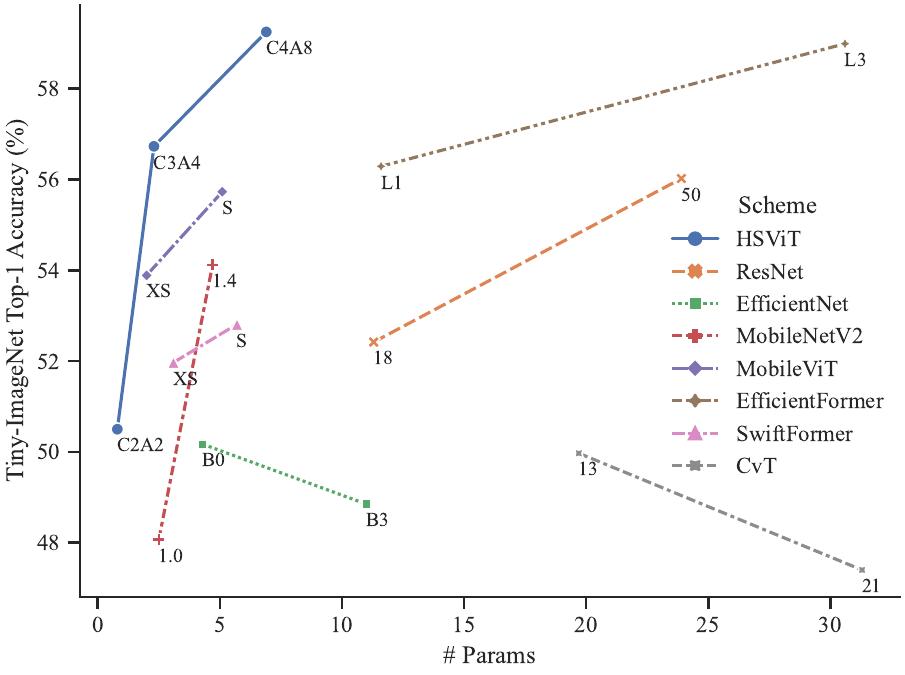}
    \caption{Number of parameters vs. Tiny-ImageNet top-1 accuracy (\%). HSViT achieves higher top-1 accuracy when solely trained and tested on Tiny-ImageNet (a small dataset) due to its better preservation of inductive bias.}
    \label{fig:num_param_vs_acc}
\end{figure}

To address the aforementioned challenges, this paper presents a novel Horizontally Scalable Vision Transformer (HSViT) scheme. First, a new image-level feature embedding for ViT is introduced, where multiple convolutional kernels operate concurrently on an image to extract their respective image-level feature maps. The feature maps are then flattened to serve as embeddings for the Transformer, enabling the self-attention mechanism to establish long-range dependencies among them. In contrast to utilizing patch-level position embeddings in classic ViT, the image-level feature embedding preserves the inductive biases in the image by allowing patterns of interest to remain in the relative positions among different embeddings through convolution and down-sampling. The image-level feature embedding effectively eliminates the pre-training requirements of ViTs while enabling its outstanding performance on small datasets, as shown in Fig.~\ref{fig:num_param_vs_acc}.

Second, a novel horizontally scalable self-attention architecture is designed to allow the model to be deployed across multiple computing devices (GPUs), and enhance performance by adding more devices. Specifically, the flattened feature maps are divided into several attention groups, with self-attention computed within each group. Multiple computing devices can independently load and calculate these feature maps and attention groups, while one of them then aggregates their predictions to produce a final forecast. The comparison between the horizontally scalable self-attention architecture with other hybrid ViT architectures is shown in Fig.~\ref{fig:arch_comparison}.

Extensive experimental results demonstrate that, without pre-training on large-scale datasets, HSViT achieves up to $10\%$ higher top-1 accuracy than the state-of-the-art CNN and ViT schemes on small datasets. Besides, the proposed HSViT can be easily transplanted to existing CNN backbones by employing them to extract image-level feature maps, improving performance by up to $3.1\%$ on ImageNet-1k. The main contributions of this paper are summarized as follows:
\begin{itemize}
    \item A novel image-level feature embedding for ViT is proposed, where the preserved inductive bias allows the model to eliminate pre-training requirements on large-scale datasets while outperforming on small datasets. 
    \item A novel horizontally scalable self-attention architecture is designed, which improves resource utilization and model scalability by enabling independent computation of submodules across computing devices.
    \item Extensive experiments are conducted, validating that HSViT effectively preserves inductive biases and supports integration with existing CNN backbones to improve their performance.
\end{itemize}

\section{Related Work}
\label{sec:related_work}

\begin{figure*}
    \centering
    \includegraphics[width=\linewidth]{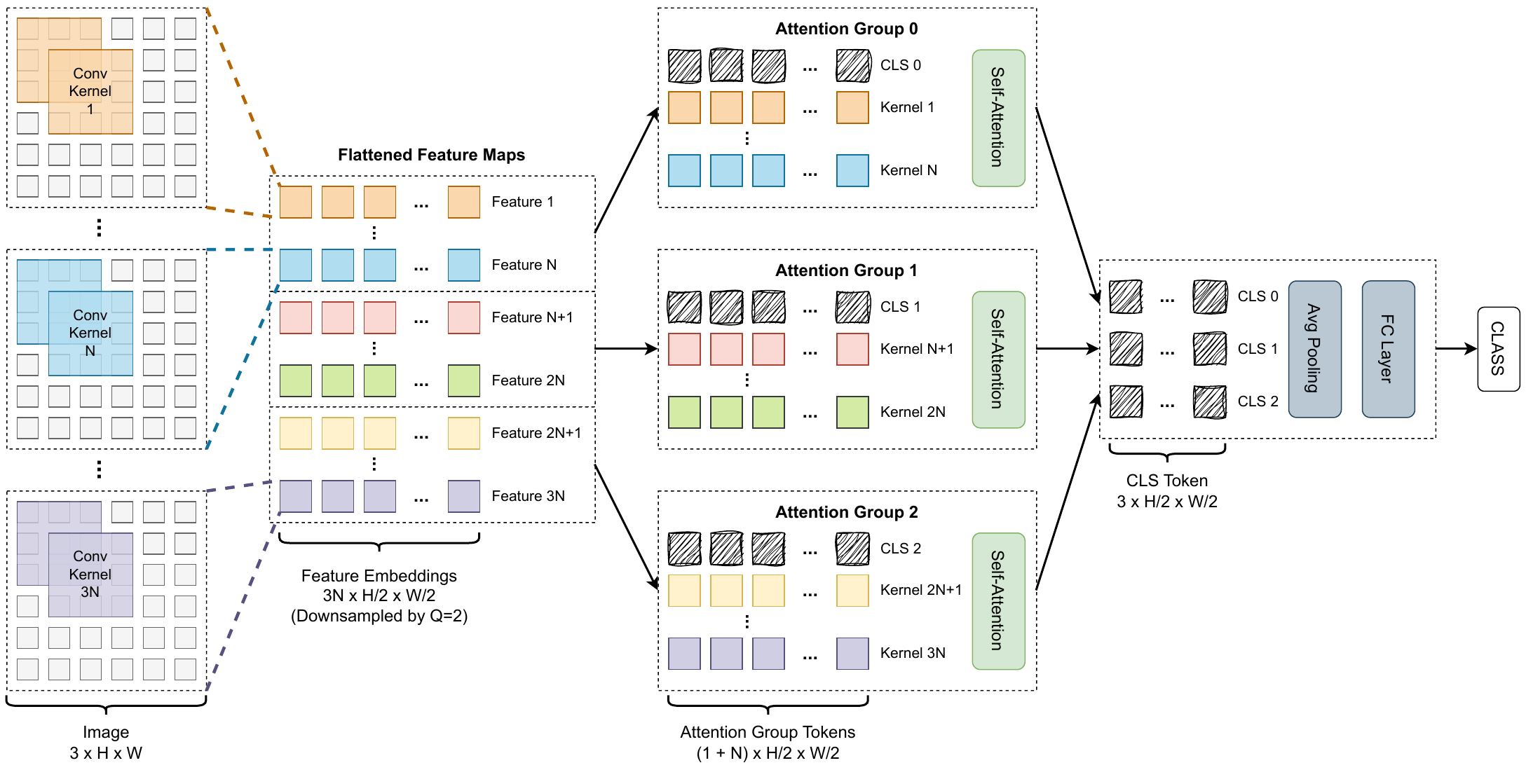}
    \caption{Feature processing pipeline of HSViT.}
    \label{fig:pipeline}
\end{figure*}

\textbf{Patch-Level Feature Embedding}: 
Transformer and its self-attention mechanism~\cite{vaswani2017attention} are introduced to computer vision after their success in natural language processing, and achieve better performance than their CNN counterparts in a variety of tasks with the help of pre-training on large-scale datasets~\cite{dosovitskiy2021image, carion2020end}. Nevertheless, as the images are converted into patch-level embedding sequences, the introduced positional embeddings disrupt the inductive bias in computer vision, like shift, scale, and rotational invariance. Numerous approaches have been proposed to address this challenge, such as convolutional embeddings~\cite{wu2021cvt, wang2022convolutional}, relative positional embeddings~\cite{wu2021rethinking, likhomanenko2021cape}, hierarchical feature extraction~\cite{liu2021swin, liu2022swin}, and guided attention~\cite{chen2024mask}. However, all these schemes adhere to the patch-level feature embedding design in ViT. By contrast, this paper proposes a novel image-level feature embedding for ViT to preserve inductive bias in convolutional layers and eliminate computing-intensive pre-training procedures.

\textbf{Hybrid Vision Transformer}:
Convolutional layers are well recognized to assist ViTs in capturing local spatial patterns and learning hierarchical representations of spatial features in images. As a result, many hybrid vision transformer architectures that integrate convolutional layers with self-attention layers have been proposed, including single-branch~\cite{carion2020end, chen2020dynamic, dosovitskiy2021image, li2022efficientformer}, dual-branch~\cite{pan2022integration, zhang2022bootstrapping}, and hierarchical~\cite{wu2021cvt, mehta2022mobilevit, guo2022cmt, shaker2023swiftformer} architectures, as shown in Fig.~\ref{fig:arch_comparison}. In particular, single-branch architectures, like DETR~\cite{carion2020end}, utilize a full-size CNN backbone to provide low-level visual features required by the subsequent self-attention layers. Dual-branch architectures mainly focuses on reusing and fusing features from self-attention and convolutional branches~\cite{pan2022integration}. Hierarchical architectures typically alternate self-attention and convolutional layers to form a mixed-hierarchical structure, aiming at fully utilizing the specialty of CNNs and Transformers, such as CvT~\cite{wu2021cvt}, CMT~\cite{guo2022cmt}, and MobileViT~\cite{mehta2022mobilevit}. Nevertheless, the aforementioned methods usually overlook scalability, especially in optimizing the utilization of multiple GPU resources in cloud environments. Conversely, this paper presents a novel horizontally scalable architecture that preserves the inductive bias in CNNs while allowing offloading submodules onto multiple computing devices.

\textbf{Distributed Machine Learning}:
Distributed machine learning typically involves training large models on large-scale datasets and distributing the workload to numerous nodes~\cite{verbraeken2020survey}. Model parallelism (MP) is a technique where different parts of a neural network are distributed across multiple computing devices to enable efficient training and inference of large models~\cite{xu2020acceleration}. However, classic MP usually causes low GPU utility and high communication overhead between GPUs. Pipeline parallelism (PP) mitigates the GPU idling problem but still has pipeline bubbles~\cite{huang2019gpipe}. Besides, there are distributed machine learning approaches, like federated learning, that aim at preserving data privacy~\cite{xu2023asynchronous}. In contrast to the aforementioned general-purpose distributed machine learning methods, this study explicitly crafts a horizontally scalable hybrid ViT architecture to facilitate collaborative model training and inference at the module level.

\section{Proposed Model}
\label{sec:model}

This section explains HSViT from the perspectives of feature processing pipeline and horizontally scalable self-attention architecture.

\subsection{Feature Processing Pipeline}

The feature processing pipeline of HSViT is shown in Fig.~\ref{fig:pipeline}. Initially, multiple convolutional kernels, each with a single output channel, are employed to extract features from the input image concurrently. The rationale behind this design is that each convolutional kernel captures one certain feature, and many of them construct the entire feature map needed for making a final prediction. As a result, convolutional kernels can be grouped and computed at different nodes to better utilize computational resources on clusters and extract as many features as possible. Assuming there are $K$ nodes in the cluster and each of them processes $N$ features, the total number of features handled by the cluster is $K \times N$. Fig.~\ref{fig:pipeline} shows the case with $K=3$. With the help of pooling layers, the feature map from each kernel is downsampled by $Q$ and flattened as an image-level feature embedding. As shown in Fig.~\ref{fig:pipeline}, after downsampling an image with size $3 \times H \times W$ by $Q=2$, the size of image-level feature embeddings becomes $3N \times H/2 \times W/2$. 

It is easy to divide image-level embeddings into non-overlapping attention groups, as each embedding contains the compressed spatial relationship of a certain feature in an image. As shown in Fig.~\ref{fig:pipeline}, the feature embeddings are divided into $3$ attention groups, with each of them having the shape of $N \times H/2 \times W/2$. Similar to the design in ViT~\cite{dosovitskiy2021image}, each attention group adds a CLS token for prediction. Therefore, each attention group possesses tokens with shape $(1+N) \times H/2 \times W/2$. After $K$ nodes have calculated self-attentions in their respective attention groups, the generated CLS tokens are aggregated by average pooling, and fed into a Fully Connected (FC) layer for final prediction. Prediction embeddings from the majority of attention groups will significantly influence the final choice if they contain strong signals indicating the class to which the image belongs.

\subsection{Horizontally Scalable Self-Attention}

\begin{figure}
    \centering
    \includegraphics[width=\linewidth]{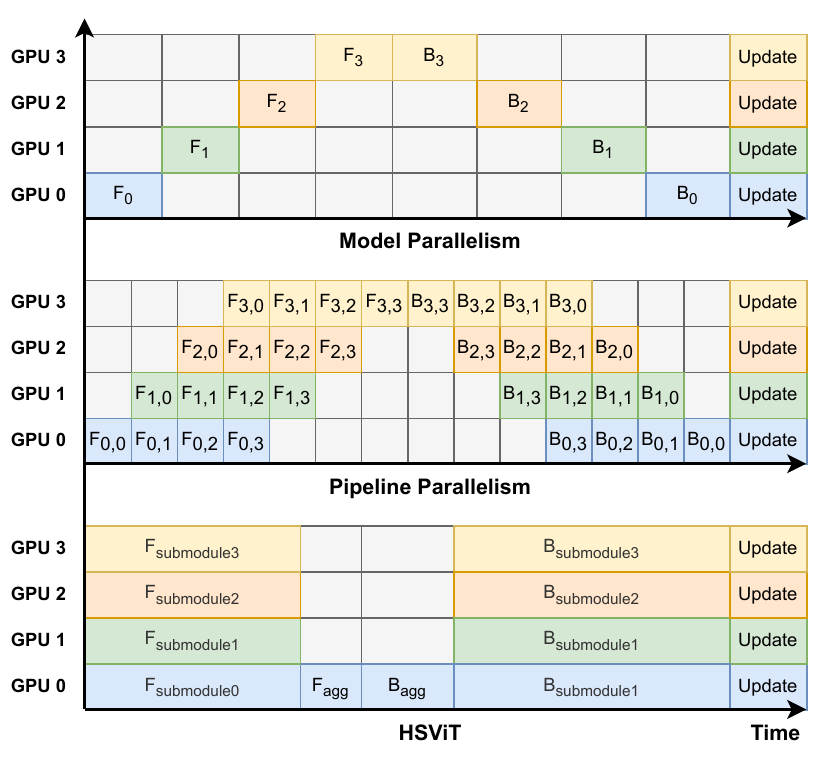}
    \caption{GPU utilization comparison between model parallelism (MP), pipeline parallelism (PP), and HSViT.}
    \label{fig:parallelism}
\end{figure}

As shown in Fig.~\ref{fig:arch_comparison}, the horizontally scalable design in HSViT is different from conventional single-branch, dual-branch, and hierarchical hybrid ViT designs by enabling submodules to compute independently. In particular, each computing device employs its own convolutional and self-attention layers to extract its feature maps and calculate the CLS token independently. These CLS tokens are then fed into one of the computing devices to make its final prediction through a voting-like mechanism. By distributing submodules across multiple GPUs, it can easily scale with the number of available GPUs.

The horizontally scalable self-attention architecture minimizes the need for frequent data transfers between GPUs, as communication occurs primarily during the aggregation of CLS tokens. Besides, by grouping convolutional kernels and processing them concurrently on different nodes, HSViT maximizes resource utilization and parallelism. In particular, Fig.~\ref{fig:parallelism} compares the GPU utilization of MP, PP, and HSViT. The idle time between computations is shown by empty cells. MP involves the sequential processing of forward pass and backpropagation for four model layers (F\textsubscript{0}, F\textsubscript{1}, F\textsubscript{2}, F\textsubscript{3}, B\textsubscript{0}, B\textsubscript{1}, B\textsubscript{2}, B\textsubscript{3}) by each GPU, which results in a large amount of idle time in between computations. For PP, input is divided into micro-batches (F\textsubscript{0,0}, F\textsubscript{0,1}, etc.) and GPUs process different stages simultaneously to reduce idle time. However, pipeline bubbles still exist at the start, middle, and end of processing. By contrast, HSViT distributes submodules (F\textsubscript{submodule0}, F\textsubscript{submodule1}, etc.) across GPUs so that they operate in parallel for most of the time. There is only minimal idle time during aggregation (F\textsubscript{agg} and B\textsubscript{agg} on GPU 0).

To fairly compare the GPU utilization, the Idle Time Ratio (ITR) is defined as the ratio of idle GPU time to the overall calculation time (without considering communication). Lower values indicate better GPU utilization. Assume the neural network has $K$ layers and is evenly assigned to $K$ same GPUs, $T_f$ and $T_b$ represent the time for forward and backpropagation operations on a minibatch or microbatch, $S$ is the number of pipeline stages in PP, and $T^\text{submodule}$ and $T^\text{agg}$ represent the computing times for a submodule and aggregation in HSViT. Consequently, the ITR for MP, PP, and HSViT can be expressed by Eq.~\ref{eq:itr_mp}, Eq.~\ref{eq:itr_pp}, and Eq.~\ref{eq:itr_hsvit}, respectively.

\begin{equation}
\label{eq:itr_mp}
\text{ITR}_\text{MP} = \frac{T_f K(K-1) + T_b K(K-1)}{T_f K + T_b K} = K-1
\end{equation}

\begin{equation}
\label{eq:itr_pp}
\text{ITR}_\text{PP} = \frac{T_f K(K-1) + T_b K(K-1)}{T_f S K + T_b S K} = \frac{K-1}{S}
\end{equation}

\begin{equation}
\label{eq:itr_hsvit}
\text{ITR}_\text{HSViT} = \frac{T_f^\text{agg} + T_b^\text{agg}}{T_f^\text{submodule}K + T_f^\text{agg} + T_b^\text{agg} + T_b^\text{submodule}K}
\end{equation}

For MP and PP, the idle time grows linearly with the number of GPUs. Compared with MP, PP reduces idle time by increasing the number of pipeline stages; however, this results in more frequent communications across GPUs. For HSViT, GPU idle time can be effectively reduced by simplifying the aggregation layer and shortening aggregation time. Besides, adding more GPUs will improve the efficiency of HSViT in utilizing GPUs, demonstrating its horizontal scalability.

HSViT offers several advantages in multi-GPU environments, including higher GPU utilization by keeping most GPUs active and minimizing idle time. It demonstrates superior scalability, enabling improved model performance by adding more GPUs. The architecture also ensures a balanced workload by evenly distributing computations across available GPUs, preventing bottlenecks.

\section{Experiments}
\label{sec:experiments}
This section presents the experiments with regard to implementation, image classification, ablation studies, sensitivity analysis, and further discussion.

\subsection{Implementation}

\begin{table}[htpb]
    \caption{Model Variants.}
    \label{table:model_variants}
    \footnotesize
    \centering
    \begin{tabular}{c|c|c|c}
    \hline
    & \multicolumn{3}{c}{Input Size} \\
    \cline{2-4}
    & $32 \times 32$ & $64 \times 64$ & $128 \times 128$ \\
    \cline{2-4}
    & \multicolumn{3}{c}{Block Design, [Output/Embedding Size]} \\
    \hline
    \multirow{3}{*}{C2A2} & CB, [$16 \times 16$] & CB, [$16 \times 16$] & CB, [$32 \times 32$] \\
    & CB, [$8 \times 8$] & CB, [$8 \times 8$] & CB, [$8 \times 8$] \\
    & $2$ $\times$ MHSA, [$64$] & $2$ $\times$ MHSA, [$64$] & $2$ $\times$ MHSA, [$64$] \\
    \hline
    \multirow{4}{*}{C3A4} & CB, [$16 \times 16$] & CB, [$32 \times 32$] & CB, [$32 \times 32$] \\
    & CB, [$16 \times 16$] & CB, [$16 \times 16$] & CB, [$16 \times 16$] \\
    & CB, [$8 \times 8$] & CB, [$8 \times 8$] & CB, [$8 \times 8$] \\
    & $4$ $\times$ MHSA, [$64$] & $4$ $\times$ MHSA, [$64$] & $4$ $\times$ MHSA, [$64$] \\
    \hline
    \multirow{5}{*}{C4A8} & CB, [$16 \times 16$] & CB, [$32 \times 32$] & CB, [$64 \times 64$] \\
    & CB, [$16 \times 16$] & CB, [$16 \times 16$] & CB, [$32 \times 32$] \\
    & CB, [$8 \times 8$] & CB, [$8 \times 8$] & CB, [$16 \times 16$] \\
    & CB, [$8 \times 8$] & CB, [$8 \times 8$] & CB, [$8 \times 8$] \\
    & $8$ $\times$ MHSA, [$64$] & $8$ $\times$ MHSA, [$64$] & $8$ $\times$ MHSA, [$64$] \\
    \hline
    \end{tabular}
    \begin{tablenotes}
      \footnotesize
      \item \textsuperscript{1}CB: Convolutional Block.
      \item \textsuperscript{2}MHSA: Multi-Head Self-Attention Block.
    \end{tablenotes}
\end{table}

\noindent\textbf{Model Variants}: The proposed HSViT scheme involves multiple hyperparameters, such as the number of convolutional kernels, convolutional layers, attention groups, convolutional layer depth, and attention layer depth. To compare with the state-of-the-art schemes, three sizes of HSViT models are designed: HSViT-C2A2, HSViT-C3A4, and HSViT-C4A8, where the number in the name indicates the number of convolutional blocks and multi-head self-attention blocks. Each convolutional block includes two Conv2d layers and one max-pooling layer. Each MHSA block includes one multi-head self-attention operation. As the convolutional layers get deeper, more convolutional kernels are required for extracting more distinct features. As a result, the number of kernels for convolutional blocks is set to $64$, $128$, $256$, and so on if there are any. For all model variants, the embedding size is set to $64$ (flattened by an $8 \times 8$ feature map) for the self-attention layers to obtain sufficient convolutional features and their relative location information. By default, the number of attention groups is set to $16$. Table~\ref{table:model_variants} depicts detailed building block designs of the variant models. HSViT is also incorporated with current CNN backbones, such as ResNet-50 and EfficientNet-B4, to confirm its wide application. These existing CNN backbones can be deployed in a distributed manner by evenly spreading convolutional kernels (also referred to as feature channels) across devices. 

\noindent\textbf{Datasets}: 
Small datasets are utilized to train models from scratch to validate the effectiveness of HSViT in retaining inductive biases, including CIFAR-10~\cite{krizhevsky09learningmultiple}, CIFAR-100~\cite{krizhevsky09learningmultiple}, Fashion-MNIST~\cite{xiao2017fashion}, Tiny-ImageNet~\cite{wu2017tiny}, and Food-101~\cite{bossard14}. For certain models that do not support small input sizes, the image size is upsampled to ensure proper training. Besides, ImageNet-1k~\cite{russakovsky2015imagenet} is used to assess the effectiveness of HSViT integration with existing CNN backbones.

\noindent\textbf{Training Details}: 
The proposed models are implemented on the PyTorch~\cite{paszke2019pytorch} framework. AdamW~\cite{loshchilov2018decoupled} is adopted as the optimizer, with the learning rate set to $0.001$ and the weight decay set to $0.01$ by default. The learning rate is adjusted through the cosine annealing method. On small datasets, models are trained from scratch for $300$ epochs. 

\subsection{Image Classification}

\begin{table*}[htpb]
    \caption{Top-1 Accuracy (\%) Comparison of Models When Trained From Scratch on Small Datasets.}
    \label{table:sota_comparison}
    \footnotesize
    \begin{tabular}{cc|c|c|c|c|c|c|c}
    \toprule
    \multicolumn{2}{c|}{\multirow{2}{*}{\textbf{Model}}} & \multirow{2}{*}{\textbf{\# Param.\footnotemark[1]}} & \multirow{2}{*}{\textbf{FLOPs\footnotemark[1]}}
    & \multicolumn{5}{c}{\textbf{Top-1 Accuracy (\%)}} \\
    \cline{5-9}
    & & & & \textbf{CIFAR-10} & \textbf{CIFAR-100} & \textbf{Fashion-MNIST} & \textbf{Tiny-ImageNet} & \textbf{Food-101} \\
    \midrule
    \multirow{6}{*}{CNN} 
    & ResNet-18~\cite{he2016deep} 
    & $11.3$ M & 0.3 G
    & $92.56$ & $69.08$ & $95.63$ & $52.42$ & $69.36$ \\
    & ResNet-50~\cite{he2016deep} 
    & $23.9$ M & 0.7 G
    & $93.69$ & $71.03$ & $\textbf{95.92}$ & $56.02$ & $74.78$ \\
    & EfficientNet-B0~\cite{tan2019efficientnet} 
    & $4.3$ M & 0.1 G
    & $90.69$ & $68.41$ & $95.26$ & $50.17$ & $73.55$ \\
    & EfficientNet-B3~\cite{tan2019efficientnet} 
    & $11.0$ M & 0.2 G
    & $90.57$ & $68.77$ & $87.17$ & $48.85$ & $73.65$ \\
    & MobileNetV2-1.0~\cite{sandler2018mobilenetv2} 
    & $2.5$ M & 0.1 G
    & $92.12$ & $66.25$ & $95.22$ & $48.07$ & $69.58$ \\
    & MobileNetV2-1.4~\cite{sandler2018mobilenetv2} 
    & $4.7$ M & 0.1 G
    & $92.62$ & $69.62$ & $95.74$ & $54.12$ & $75.85$ \\
    \hline
    ViT 
    & ViT-B/16\footnotemark[2]~\cite{dosovitskiy2021image} 
    & $85.4$ M & 46.1 G
    & $83.33$ & $62.10$ & $93.55$ & $55.51$ & $73.64$ \\
    \hline
    \multirow{8}{*}{Hybrid} & MobileViT-XS~\cite{mehta2022mobilevit} 
    & $2.0$ M & 0.1 G
    & $93.09$ & $70.83$ & $95.53$ & $53.89$ & $75.45$ \\
    & MobileViT-S~\cite{mehta2022mobilevit} 
    & $5.1$ M & 0.2 G
    & $92.87$ & $70.97$ & $95.73$ & $55.73$ & $68.00$ \\
    & EfficientFormer-L1~\cite{li2022efficientformer} 
    & $11.6$ M & 0.2 G
    & $92.57$ & $72.79$ & $95.87$ & $56.29$ & $75.28$ \\
    & EfficientFormer-L3~\cite{li2022efficientformer} 
    & $30.6$ M & 0.6 G
    & $93.34$ & $73.82$ & $95.71$ & $58.99$ & $77.21$ \\
    & SwiftFormer-XS~\cite{shaker2023swiftformer} 
    & $3.1$ M & 0.1 G
    & $92.34$ & $68.19$ & $95.61$ & $51.97$ & $69.82$ \\
    & SwiftFormer-S~\cite{shaker2023swiftformer} 
    & $5.7$ M & 0.2 G
    & $92.56$ & $68.23$ & $95.87$ & $52.81$ & $71.12$ \\
    & CvT-13~\cite{wu2021cvt} 
    & $19.7$ M & 1.0 G
    & $87.91$ & $63.32$ & $94.76$ & $49.97$ & $61.44$\footnotemark[2] \\
    & CvT-21~\cite{wu2021cvt} 
    & $31.3$ M & 1.7 G
    & $87.65$ & $61.93$ & $94.89$ & $47.40$\footnotemark[2] & $62.21$ \\
    \hline
    \multirow{3}{*}{Ours} 
    & HSViT-C2A2 
    & $0.8$ M & 0.4 G
    & $90.64$ & $67.84$ & $95.10$ & $50.50$ & $67.88$ \\
    & HSViT-C3A4 
    & $2.3$ M & 1.3 G
    & $93.04$ & $72.46$ & $95.72$ & $56.73$ & $73.54$ \\
    & HSViT-C4A8 & 
    $6.9$ M & 1.9 G
    & $\textbf{94.04}$ & $\textbf{73.85}$ & $\textbf{95.92}$ & $\textbf{59.25}$ & $\textbf{79.06}$ \\
    \bottomrule
    \end{tabular}
    \begin{tablenotes}
      \footnotesize
      \item \textsuperscript{1}Calculated when training models on Tiny-ImageNet with image size $64 \times 64$.
      \item \textsuperscript{2}The learning rate is adjusted to $1 \times 10^{-4}$ for convergence.
    \end{tablenotes}
\end{table*}

As shown in Table~\ref{table:sota_comparison}, HSViT achieves higher top-1 accuracy than the state-of-the-art schemes with similar parameter numbers due to its better preservation of inductive bias. For example, on Tiny-ImageNet, HSViT-C3A4 achieves $56.73\%$ top-1 accuracy with $2.3$ M parameters, surpassing SwiftFormer-XS by $4.76\%$ ($3.1$ M, $51.97\%$), MobileViT-XS by $2.84\%$ ($2.0$ M, $53.89\%$), and MobileNetV2-1.0 by $8.66\%$ ($2.5$ M, $48.07\%$). With $6.9$ M parameters, HSViT-C4A8 achieves the highest top-1 accuracy among the five datasets, except for Fashion-MNIST, where it shares the same $95.92\%$ accuracy as ResNet-50. The relatively high FLOPS (Floating Point Operations Per Second) is due to the pile of convolutional layers with large feature channels (kernels), which can be mitigated through depthwise convolution~\cite{sandler2018mobilenetv2}.

It can be observed that certain schemes show a decreased top-1 accuracy after increasing the model size. For instance, MobileViT-S ($5.1$ M, $68.00\%$) has a lower top-1 accuracy than MobileViT-XS ($2.0$ M, $75.45\%$) on Food-101. EfficientNet-B3 ($11.0$ M, $48.85\%$) is inferior to EfficientNet-B0 ($4.3$ M, $50.17\%$) on Tiny-ImageNet. This phenomenon occurs because the deeper layers or larger embeddings in the models present additional challenges to training, especially on small datasets. By contrast, the reserved inductive bias in HSViT allows it to be trained easily, while the architecture enables better horizontal scalability across multiple computing devices. 

\begin{table}[htpb]
\caption{Top-1 Accuracy (\%) Comparison of Models on ImageNet-1k With HSViT Integrating with Existing CNN Backbones.}
\label{table:backbone_imagenet}
\footnotesize
\begin{tabular}{cc|c|c|c}
\toprule
\multicolumn{2}{c|}{\textbf{Model}}  & \textbf{\# Param.} & \textbf{FLOPs} & \textbf{Acc.} \\
\midrule
\multirow{5}{*}{CNN}
& MobileNetV2-1.4~\cite{sandler2018mobilenetv2}
& 4.3 M   & 0.6 G  & 74.7 \\
& ResNet-50~\cite{he2016deep}
& 24.4 M  & 4.1 G  & 76.1 \\
& ResNet-101~\cite{he2016deep}
& 42.5 M  & 7.8 G  & 77.4 \\
& EfficientNet-B4~\cite{tan2019efficientnet}
& 17.5 M  & 8.8 G  & 82.5 \\
& EfficientNet-B5~\cite{tan2019efficientnet}
& 30.0 M  & 9.9 G  & 83.3 \\
\hline
\multirow{3}{*}{ViT}
& ViT-B/16~\cite{dosovitskiy2021image}
& 86.0 M  & 55.5 G  & 77.9 \\
& ViT-L/16~\cite{dosovitskiy2021image}
& 307.0 M & 191.1 G & 76.5 \\
& EfficientViT-M5~\cite{liu2023efficientvit}
& 12.4 M  & 0.5 G   & 77.1 \\
\hline
\multirow{10}{*}{Hybrid}
& EfficientFormer-L1~\cite{li2022efficientformer}
& 12.3 M  & 1.3 G  & 79.2 \\
& EfficientFormer-L3~\cite{li2022efficientformer}
& 31.3 M  & 3.9 G  & 82.4 \\
& SwiftFormer-L1~\cite{shaker2023swiftformer}
& 12.1 M  & 1.6 G  & 80.9 \\
& SwiftFormer-L3~\cite{shaker2023swiftformer}
& 28.5 M  & 4.0 G  & 83.0 \\
& CvT-13~\cite{wu2021cvt}
& 19.7 M  & 4.5 G  & 81.6 \\
& CvT-21~\cite{wu2021cvt}
& 31.3 M  & 7.1 G  & 82.5 \\
& FastViT-SA24~\cite{vasu2023fastvit}
& 20.6 M  & 3.8 G  & 82.6 \\
& FastViT-SA36~\cite{vasu2023fastvit}
& 30.4 M  & 5.6 G  & 83.6 \\
& SMT-T~\cite{lin2023scale}
& 11.5 M  & 2.4 G  & 82.2 \\
& SMT-S~\cite{lin2023scale}
& 20.5 M  & 4.7 G  & 83.7 \\
\hline
\multicolumn{2}{c|}{HSViT + MobileNetV2-1.4} & 6.1 M  & 1.3 G  & 54.9 \\
\multicolumn{2}{c|}{HSViT + ResNet-50}  & 25.3 M & 8.3 G  & 78.0 \\
\multicolumn{2}{c|}{HSViT + ResNet-101} & 44.3 M & 15.7 G & 80.5 \\
\multicolumn{2}{c|}{HSViT + EfficientNet-B4} & 28.7 M & 9.5 G  & 82.7 \\
\multicolumn{2}{c|}{HSViT + EfficientNet-B5} & 56.9 M & 22.7 G & 83.4 \\
\bottomrule
\end{tabular}
\end{table}

Integrating HSViT with various CNN backbones has demonstrated notable performance improvements, with enhancements of up to $3.1\%$ on ImageNet-1k compared to using the existing CNN backbone alone, as shown in Table~\ref{table:backbone_imagenet}. For instance, the integration with ResNet-50 increased the top-1 accuracy from $76.1\%$ to $78.0\%$, while integrating with EfficientNet-B4 improved the accuracy from $82.5\%$ to $82.7\%$. Although this integration results in an increase in parameter numbers and FLOPs, this overhead can be mitigated by offloading submodules onto multiple computing devices, optimizing resource utilization. These results validate the broad applicability of HSViT to various CNN backbones. However, in the case of HSViT + MobileNetV2-1.4, the top-1 accuracy decreased significantly to $54.9\%$. This decline is likely due to the use of unified training parameters without fine-tuning specific to the MobileNetV2-1.4 architecture, as well as its relatively smaller number of convolution kernels ($28$ compared to $32$ in ResNet-50) for each attention group.

\subsection{Ablation Studies}

\begin{table}[htpb]
    \caption{Ablation Study on CIFAR-10 and CIFAR-100.}
    \label{table:ablation_study}
    \footnotesize
    \centering
    \begin{tabular}{c|c|c|c|c}
    \toprule
    \multirow{2}{*}{\textbf{Model}} & \multicolumn{2}{c|}{\textbf{Layer}} &
    \multicolumn{2}{c}{\textbf{Top-1 Accuracy (\%)}} \\
    \cline{2-5}
    & \textbf{Attn} & \textbf{Conv2d} & \textbf{CIFAR-10} & \textbf{CIFAR-100} \\
    \midrule
    \multirow{3}{*}{HSViT-C2A2} & \checkmark & & $60.43$ & $32.29$ \\
    & & \checkmark & $87.36$ & $37.92$ \\
    & \checkmark & \checkmark & $\textbf{90.64}$ & $\textbf{67.84}$ \\
    \hline
    \multirow{3}{*}{HSViT-C3A4} & \checkmark & & $63.52$ & $35.17$ \\
    & & \checkmark & $91.87$ & $42.83$ \\
    & \checkmark & \checkmark & $\textbf{93.04}$ & $\textbf{72.46}$ \\
    \hline
    \multirow{3}{*}{HSViT-C4A8} & \checkmark & & $64.73$ & $37.29$ \\
    & & \checkmark & $93.32$ & $43.51$ \\
    & \checkmark & \checkmark & $\textbf{94.04}$ & $\textbf{73.85}$ \\
    \bottomrule
    \end{tabular}
    \begin{tablenotes}
      \footnotesize
      \item \textsuperscript{1}Layers without the ``\checkmark'' mark are ablated.
    \end{tablenotes}
\end{table}

\noindent\textbf{Ablate Self-Attention Layer}: On CIFAR-10 and CIFAR-100, the top-1 accuracy of HSViT declines with the ablation of the self-attention layers. In particular, as shown in Table~\ref{table:ablation_study}, on CIFAR-10, as the number of convolutional blocks grows (from C2 to C4), the ablation of the self-attention layer has a decreasing effect on the top-1 accuracy, from $3.28\%$ to $0.72\%$. However, the phenomenon is not as pronounced on CIFAR-100. Although C2 and C4 have different numbers of convolutional blocks, the ablation of the self-attention layer consistently results in a significant reduction in top-1 accuracy, approximately $30\%$. The reason is that C4 alone offers enough convolutional layers to extract and aggregate features from CIFAR-10, resulting in an acceptable top-1 accuracy of $93.32\%$. However, C4 alone has insufficient feature aggregation capability and understanding in handling CIFAR-100, where the number of classes is increased tenfold while the number of samples for each class is reduced to a tenth. Comparing the absolute drop in top-1 accuracy between CIFAR-10 and CIFAR-100 after ablating the self-attention layer further supports this insight. Take HSViT-C2A2 as an example, the reduction is $3.28\%$ for CIFAR-10 but is $29.92\%$ for CIFAR-100, indicating that the self-attention layer imparts a richer comprehension of the CIFAR-100 features to HSViT.

\noindent\textbf{Ablate Convolutional Layer}:
Ablating the convolutional layer from HSViT reveals a significant drop in top-1 accuracy of approximately $30\%$ and $36\%$ for CIFAR-10 and CIFAR-100, respectively. This drop demonstrates that the image-level convolutional kernel is crucial for extracting useful features for the subsequent self-attention layers. In addition, the convolutional layer consistently has a more significant impact on top-1 accuracy than the self-attention layer. For example, for HSViT-C4A8 on CIFAR-10, when ablating the convolutional and self-attention layers, the top-1 accuracy reduction is $29.31\%$ vs. $0.72\%$. This phenomenon occurs because the self-attention layer alone struggles to learn local features with insufficient data, making it highly dependent on the convolutional layers for local information extraction, as noted in the literature~\cite{raghu2021vision}. Nevertheless, the effectiveness of the self-attention layer remains significant, as it offers a useful understanding of various features and improves top-1 accuracy with relatively few layers of parameters, especially when the classification task becomes more complicated, like CIFAR-100.

\subsection{Sensitivity Analysis}

\begin{figure}
    \centering
    \includegraphics[width=\linewidth]{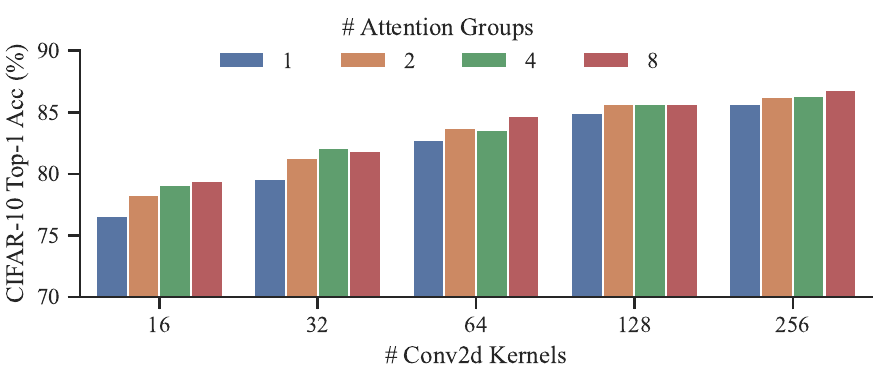}
    \caption{Number of convolutional kernels vs. number of attention groups. \textmd{As the number of convolutional kernels and attention groups grows, the top-1 accuracy on CIFAR-10 rises.}}
    \label{fig:num_conv_kernel_vs_num_attn}
\end{figure}

\begin{figure}
    \centering
    \includegraphics[width=\linewidth]{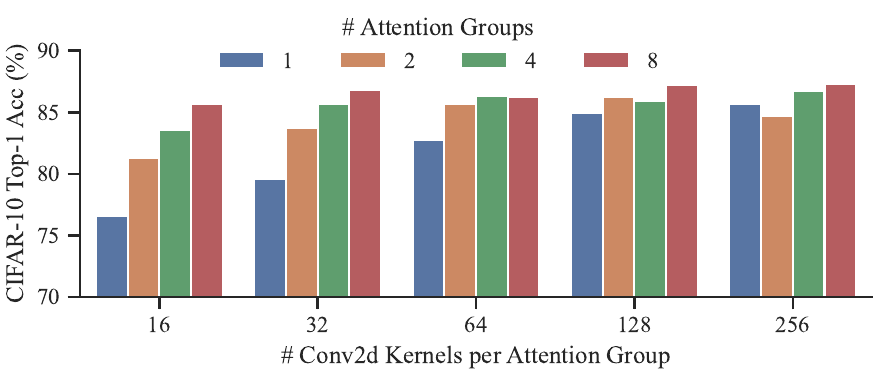}
    \caption{Number of convolutional kernels per attention group vs. number of attention groups. The top-1 accuracy on CIFAR-10 plateaus once the model has learned a sufficient number of features, regardless of the number of convolutional kernels or attention groups.}
    \label{fig:fix_num_conv_kernel_per_attn}
\end{figure}

\begin{figure}
    \centering
    \includegraphics[width=\linewidth]{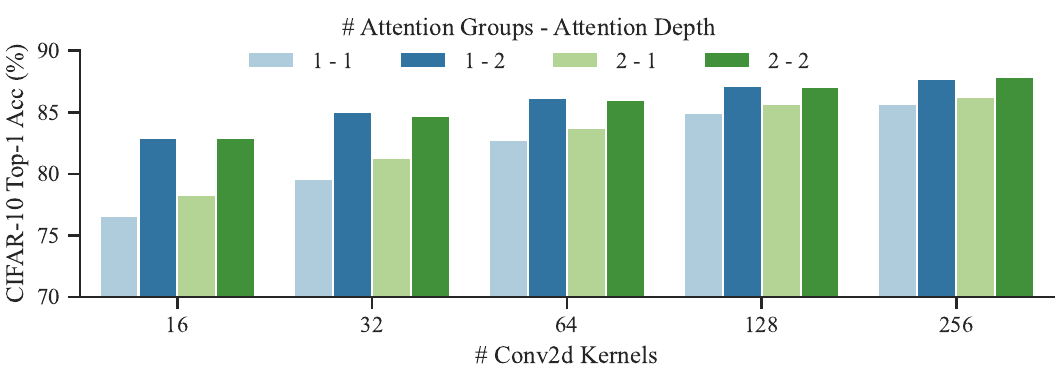}
    \caption{Number of attention groups and attention depth. A deeper self-attention layer compensates for the reduced number of attention groups in feature understanding.}
    \label{fig:attn_group_vs_attn_depth}
\end{figure}

\begin{figure}
    \centering
    \includegraphics[width=\linewidth]{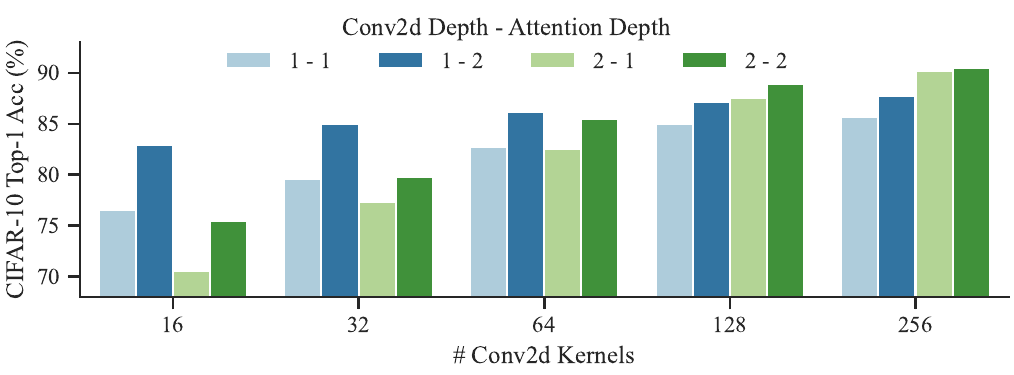}
    \caption{Convolution depth and attention depth. More convolutional kernels necessitate both deeper convolutional and deeper self-attention layers to achieve optimal accuracy.}
    \label{fig:conv_depth_vs_attn_depth}
\end{figure}

\noindent\textbf{Convolutional Kernels and Attention Groups}: 
As shown in Fig.~\ref{fig:num_conv_kernel_vs_num_attn}, the top-1 accuracy on CIFAR-10 increases with the number of convolutional kernels and attention groups. This observation aligns with intuition: a higher number of convolutional kernels enables the extraction of more features from the convolutional layers, thereby improving the discrimination of local details in images. Besides, an increased number of attention groups enhances the understanding of these extracted features and provides more voting stands during the decision stage, leading to more accurate final predictions. Nonetheless, this improvement in top-1 accuracy plateaus once the number of convolutional kernels and attention groups surpasses a certain threshold, as the model reaches its capacity for learning features given its depth. For instance, as shown in Fig.~\ref{fig:fix_num_conv_kernel_per_attn}, top-1 accuracy rises predictably as the number of convolutional kernels per attention group rises from $16$ to $32$, but the growth becomes marginal beyond $64$. Likewise, when increasing the attention groups, the accuracy gain becomes inconsistent beyond $64$ convolutional kernels per attention group and plateaus at around $87\%$.

\noindent\textbf{Attention Depth and Convolution Depth}: Fig.~\ref{fig:attn_group_vs_attn_depth} demonstrates that a deeper attention layer consistently improves top-1 accuracy, despite the improvement diminishes as the number of convolutional kernels increases. This is because network depth plays a crucial role in integrating and enriching different levels of features, as evidenced in previous research~\cite{he2016deep}. Additionally, Fig.~\ref{fig:attn_group_vs_attn_depth} shows that the number of attention groups has minimal effect on top-1 accuracy once the self-attention layers are increased to two. For example, with $64$ convolutional kernels and a single self-attention layer, the accuracy difference between one and two attention groups is approximately $1\%$. However, with two self-attention layers, this difference decreases to $0.15\%$. This indicates that deeper self-attention layers help to understand the image-level convolutional features and even compensate for the decrease in attention groups. Furthermore, Fig.~\ref{fig:conv_depth_vs_attn_depth} highlights the distinct impacts of convolution and attention depth on top-1 accuracy. In particular, the deeper attention layer always improves top-1 accuracy, while the deeper convolution layer only yields a positive effect when there is a sufficient number of convolutional kernels. This is because the first layer has half as many convolutional kernels as the second layer. When the number of convolutional kernels is relatively small, the inadequate feature representation capacity in the first layer hinders the second layer from delivering enough features for final prediction. Fig.~\ref{fig:conv_depth_vs_attn_depth} illustrates that the combination of deeper convolutional layers, deeper self-attention layers, and a sufficient number of convolutional kernels collectively enhances the performance of HSViT.

\subsection{Further Discussion}

\textbf{Transient GPU Communication Burden}: While adding more GPUs will improve the efficiency of HSViT in utilizing GPUs, it may pose an additional transient communication burden to GPUs during aggregation. To mitigate the transient GPU communication burden, several approaches can be employed, such as gradient compression~\cite{deng2020model, abrahamyan2021learned}, gradient sparsification~\cite{shi2020communication}, taking advantage of faster links between certain GPUs, overlapping the aggregation process with the computation of the subsequent batch, etc.

\textbf{High FLOPs}: Large numbers of convolutional kernels are observed to cause high FLOPs. However, adopting depthwise separable convolutions, as implemented in MobileNet~\cite{sandler2018mobilenetv2}, presents a promising approach to reduce both parameter number and FLOPs. Additionally, the horizontally scalable design of HSViT enables distributed deployment across multiple computing devices, which can effectively mitigate the computational challenges associated with high FLOPs.

\textbf{Small-Object Detection from Large Image}: The flattened image-level feature map leads to a quadratic increase in computational complexity, thereby presenting a significant challenge to identifying small objects within large, high-resolution images. A viable solution is crowdsourcing, where large images are segmented into smaller sections and distributed across multiple attention groups. However, this approach may lead to a decline in model performance.

\section{Conclusion}
\label{sec:conclusion}
This paper introduces a horizontally scalable vision transformer (HSViT) scheme. 
With a novel image-level feature embedding design, HSViT effectively preserves the inductive bias and eliminates the pre-training requirements. Besides, a novel horizontally scalable architecture is designed, enabling collaborative model training and inference across multiple computing devices. Extensive experimental results demonstrate that, without pre-training, HSViT outperforms state-of-the-art models on small datasets, affirming its superior preservation of inductive bias. Furthermore, HSViT is easily applied to existing CNN backbones, which shows up to $3.1\%$ improvement in top-1 accuracy on ImageNet-1k, demonstrating its broad applicability. Future work encompasses expanding HSViT to other computer vision tasks, mathematically and empirically analyzing communication overhead, and investigating its performance in detecting small objects.

{\small
\bibliographystyle{ieee_fullname}
\bibliography{ref}
}

\end{document}